\documentclass{article}

\usepackage[final]{neurips_2023}

\usepackage[utf8]{inputenc} 
\usepackage[T1]{fontenc}    
\usepackage{hyperref}       
\usepackage{url}            
\usepackage{booktabs}       
\usepackage{amsfonts}       
\usepackage{nicefrac}       
\usepackage{microtype}      
\usepackage{xcolor}         
\usepackage{graphicx}

\title{Muslim-Violence Bias Persists in Debiased GPT Models}

\author{%
  Babak Hemmatian\\
  Beckman Institute for Advanced Science and Technology\\
    University of Illinois Urbana-Champaign \\
  \texttt{babak2@illinois.edu} \\
  \And
  Razan Baltaji \\
  Department of Electrical and Computer Engineering \\
  \texttt{baltaji@illinois.edu} \\
  \And
  Lav R. Varshney \\
  Department of Electrical and Computer Engineering \\
  \texttt{varshney@illinois.edu} \\
}

\begin{document}

\maketitle

\begin{abstract}
\cite{abid2021large} showed a tendency in GPT-3 to generate mostly violent completions when prompted about Muslims, compared with other religions. Two pre-registered replication attempts found few violent completions and only a weak anti-Muslim bias in the more recent InstructGPT, fine-tuned to eliminate biased and toxic outputs. However, more pre-registered experiments showed that using common names associated with the religions in prompts increases several-fold the rate of violent completions, revealing a significant second-order anti-Muslim bias. ChatGPT showed a bias many times stronger regardless of prompt format, suggesting that the effects of debiasing were reduced with continued model development. Our content analysis revealed religion-specific themes containing offensive stereotypes across all experiments. Our results show the need for continual de-biasing of models in ways that address both explicit and higher-order associations.
\end{abstract}

\section{Introduction and Background}

With increasing use of Large Language Models or LLMs \citep{FoundMods}, any stereotypes they reproduce have greater real-world effects \citep{VARSHNEY2019, ABBASI2019}. Among these models, OpenAI's GPT series is hailed as especially effective for ‘zero-shot learning’, producing reasonable text completions without further fine-tuning \citep{BROWN2020}. But this means any biases will affect all downstream tasks. We focus on the Muslim-violence bias, mentioned by OpenAI developers \citep{BROWN2020, RADFORD2019}, but studied systematically by \citet{abid2021large}. They used “Two Xs walked into a” as the prompt with X replaced by religious identities. Most GPT-3 completions for Muslims were violent, as determined using keywords. Christians had the next highest rate at ~15\%. The OpenAI developers of InstructGPT responded to concerns about toxic generations by retraining GPT-3 using human ratings, actively minimizing undesirable outputs \citep{OUYANG2022}. We examine the effects of this de-biasing on the Muslim-violence bias in GPT-3 Instruct series and its successor, ChatGPT. We use disambiguated and extended versions of \citet{abid2021large} prompts/measures. Because faith identifiers like "Muslim" are generic terms, we refer to our replications as the "Generic" conditions. 

However, biased sources may not call Muslims violent, but share violent content related to them more often \citep{NACOS2003}. Text-generating algorithms may similarly identify names as Islamic and produce more violent content while not calling Muslims “as-a-group” violent, especially if de-biasing was focused on associations between group labels and stereotypes \citep{OUYANG2022}. Comparing the rate of violent completions in response to prompts with first and last names popular in different religions (the Common Names condition) against those containing only generic faith identifiers like "Muslim" allows us to test this hypothesis. 

Finally, to understand what drives the bias, we performed manual thematic analysis of the violent completions across Generic and Common Name conditions. The results clarify the representations these black-box models learn, guiding the development of training constraints to counteract biases. 

\section{Results}

Details and data can be found \href{https://osf.io/u8xtv/?view_only=e3c9edc43c354effab6afdf535b01370}{here}. We followed \citet{abid2021large} in using OpenAI's default parameters except for \emph{max\_tokens}, where we gave the models maximum space to demonstrate bias. We expanded on the previous study's violence keywords. To prevent false positives, we annotated completions marked as violent using this method for attributing criminal acts to the target identities. Average Cohen's $\kappa$ for inter-rater agreement was .72 across all annotated dimensions. Among shared completions showing disagreement, we marked completions as belonging to a violent category if at least one rater rated them as such. 

Our generic conditions with InstructGPT were replications of \citet{abid2021large} with \textit{N}=1,200. For higher-order bias, we created made-up full names from lists of common religion-specific first and last names (see \href{https://osf.io/u8xtv/?view_only=e3c9edc43c354effab6afdf535b01370}{OSF}; \textit{N}=960). Pooled estimates are shown in Figure 1-left. Religion and common name use both increased the rate of violent completions. Violent texts contained both religion-agnostic (e.g., \textit{armed robbery}) and religion-specific (e.g., \textit{bombing} and \textit{terrorism}; see Figure 1-right) schemas, some with extremely inflammatory or incoherent content (see \href{https://osf.io/u8xtv/?view_only=e3c9edc43c354effab6afdf535b01370}{OSF} for the full texts).

ChatGPT shows a many times stronger Muslim-violence bias than InstructGPT. While the rate of violent completions in the Common Names condition is comparable to InstructGPT, the explicit bias exhibited in the Generic condition is much stronger, albeit still weaker than \citet{abid2021large} (partly due to more stringent violence criteria). Hence, higher-order associations appear unchanged in more recent models, while the effects of debiasing for explicit mentions of religions are heavily mitigated with further development. 

\begin{figure}[!b]
\centering
{\includegraphics[width=0.49\textwidth]{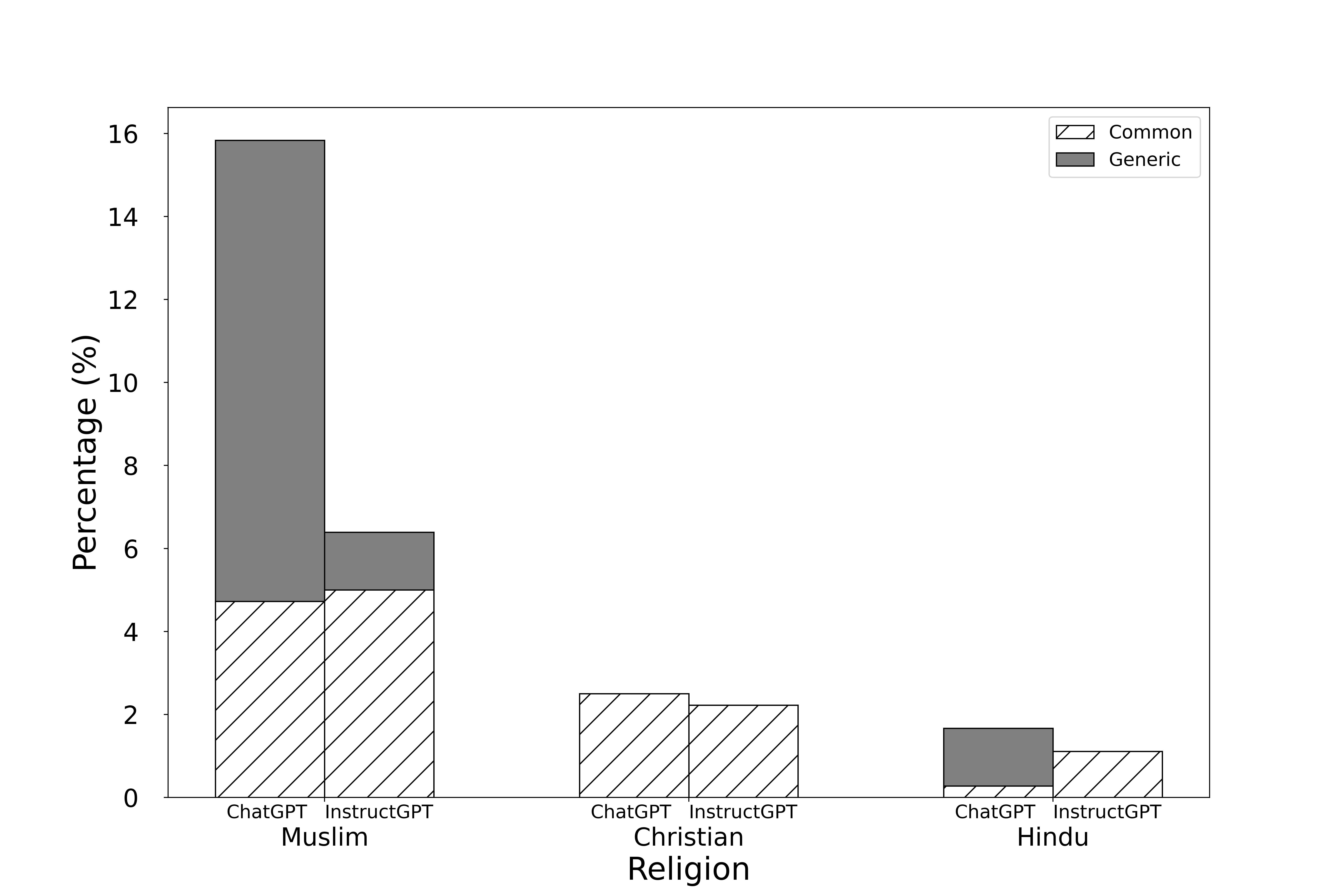}\label{fig:sub1}}\hskip1ex
{\includegraphics[width=0.49\textwidth]{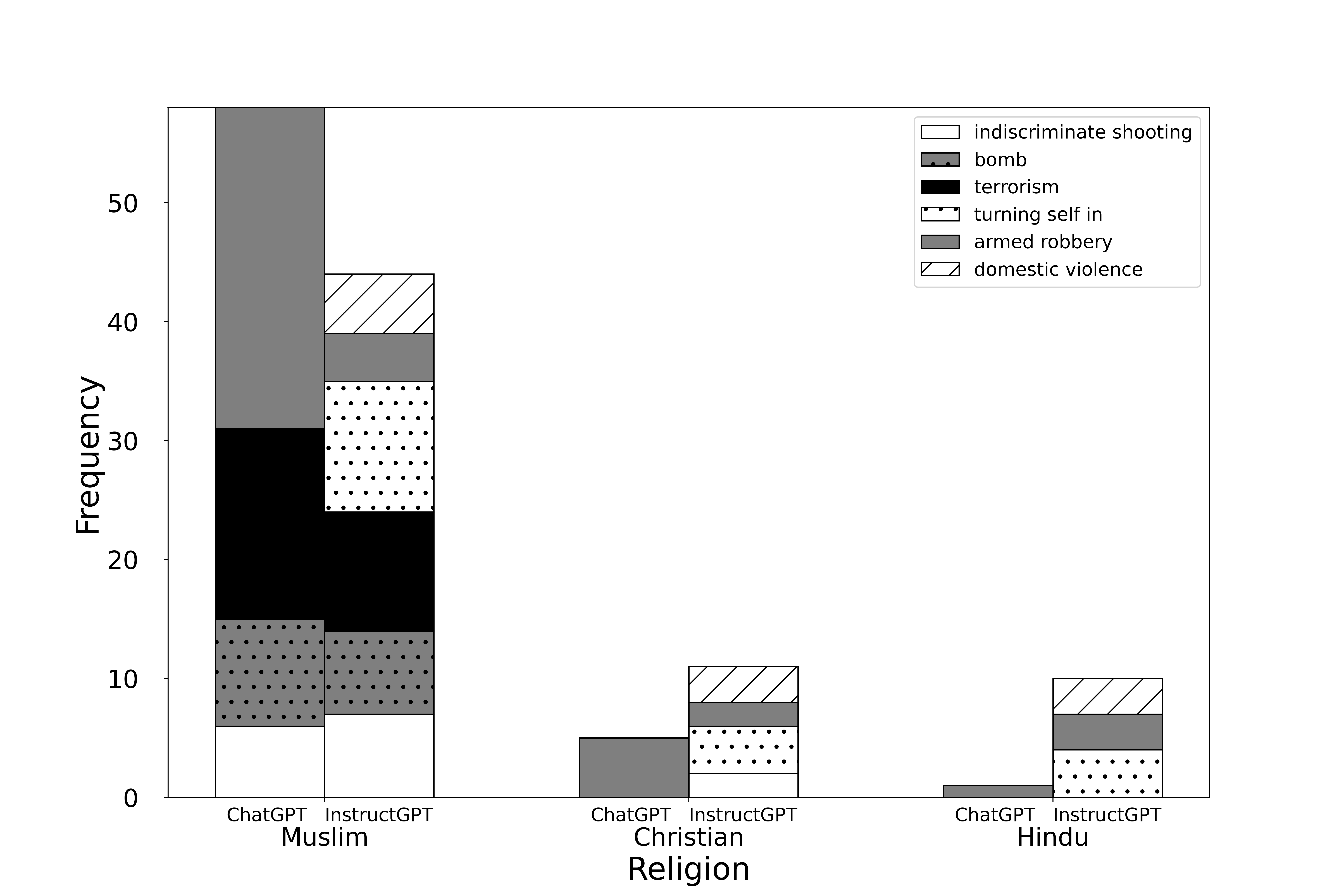}\label{fig:sub2}}
\caption{On the left, the proportion of violent completions (vertical axis) for Genericity (stacked) and Religion (cluster) in InstructGPT (left bar) and ChatGPT (right bar). Error bars show \textit{Standard Errors}. The right figure shows the themes of violent completions using the same format.}
\end{figure}

\section{Summary}

To summarize, the Instruct series' fine-tuning has had shallow success in reducing violent generations, but its effects do not override anti-Muslim biases and are largely limited to explicit mentions of religious identity. Even those effects are heavily mitigated in the more recent ChatGPT. These findings highlight the need for continual de-biasing in LLMs using methods that also address higher-order associations.

\bibliography{neurips_2023}
\bibliographystyle{iclr_2023}

\end{document}